\documentclass[sigconf,screen]{acmart}
\AtBeginDocument{%
  }

\setcopyright{acmlicensed}
\copyrightyear{2018}
\acmYear{2018}
\acmDOI{XXXXXXX.XXXXXXX}
\acmConference[FSE '26]{Companion Proceedings of the 34th ACM Symposium on the Foundations of Software Engineering}{June 5--9, 2026}{Montreal, Canada}
\acmISBN{978-1-4503-XXXX-X/2018/06}

\begin{document}

\title{TokenScope: Token-Level Explainability and Interpretability for Code-Oriented Tasks in Large Language Models}

\author{Amirreza Esmaeili}
\email{a.esmaeili@ubc.ca}
\orcid{0009-0005-5459-8559}
\affiliation{%
  \institution{University of British Columbia}
  \city{Kelowna}
  \state{British Columbia}
  \country{Canada}
}

\author{Fatemeh Fard}
\email{fatemeh.fard@ubc.ca}
\orcid{0000-0002-4505-6257}
\affiliation{%
  \institution{University of British Columbia}
  \city{Kelowna}
  \state{British Columbia}
  \country{Canada}
}

\renewcommand{\shortauthors}{Esmaeili et al.}

\begin{abstract}
    Understanding how Large Language Models (LLMs) make token-level decisions during code generation remains a major challenge for both researchers and practitioners. While recent tools provide insights into model internals or generation outcomes, they often lack decoding-time signals, fine-grained uncertainty measures, and interactive mechanisms for exploring alternative generation paths. We present TokenScope, an interactive interpretability and analysis tool for decoder-based LLMs that exposes token-level metrics, attention patterns, and structural information during generation. TokenScope supports interactive token replacement, counterfactual branching, and code-aware aggregation via abstract syntax trees. By unifying decoding-time signals with structural program analysis, TokenScope enables systematic investigation of LLM behaviour during code generation.
\end{abstract}



\received{20 February 2007}
\received[revised]{12 March 2009}
\received[accepted]{5 June 2009}

\maketitle

\section{Introduction}

Large Language Models (LLMs) based on transformer decoders have become central to modern software engineering workflows, supporting tasks such as code generation, refactoring, test synthesis, and program understanding \cite{codex-chen2021evaluating, alpha-code-li2022competition, codellama-roziere2023code}. Despite their effectiveness, these models remain opaque: practitioners and researchers are often unable to explain \emph{why} a model produced a particular token, \emph{how} uncertain it was at different points in the generation, or \emph{what alternatives} were plausibly considered but discarded. This lack of explainability is particularly problematic in code generation, where small token-level deviations can cascade into syntactic errors, semantic bugs, or incorrect API usage.

\begin{figure*}[h]
  \centering
  \includegraphics[width=\linewidth]{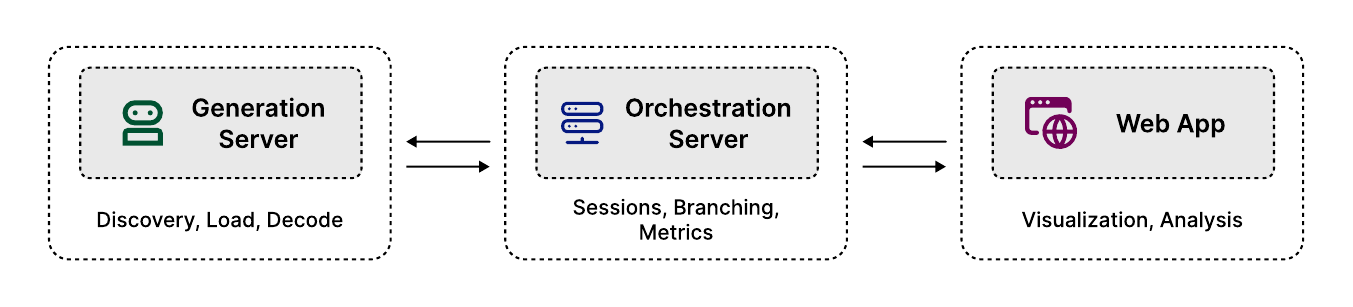}
  \caption{Overview of TokenScope’s modular architecture. The generation server exposes decoder-only LLMs, the orchestration server orchestrates generation, metrics, and branching, and the web-based frontend application provides interactive exploration of decoding-time signals.}
  \label{fig:system}
\end{figure*}

Existing interpretability and explainability tools provide partial insight into LLM behaviour but leave important gaps. Prior work largely emphasizes dataset-level analysis, post-hoc evaluation, internal representation inspection, or static attention visualization, often requiring offline analysis and substantial expertise \cite{lit-tenney2020language, nanda2022transformerlens, cooney2023SparseAutoencoder, alammar-2021-ecco, bertvis-clark2019what}. These approaches rarely expose decoding-time uncertainty, alternative token candidates, or decision margins, and they typically treat generation as a single outcome rather than as a sequence of interdependent token-level decisions. As a result, researchers lack the infrastructure to study where uncertainty arises during generation, how attention and probability mass shift over time, and how local decisions propagate into structural errors in code.

To address these challenges, we introduce \textbf{TokenScope}, an interactive explainability and interpretability environment for decoder-based LLMs, with a focus on code generation. TokenScope instruments the decoding process itself, capturing fine-grained token-level signals, including confidence measures, margin confidence, entropy, token surprisal, alternative candidates, and attention weights, at each generation step. These signals are streamed to an interactive frontend that enables direct inspection and comparison of token-level decisions.

TokenScope supports both researchers and practitioners by enabling interactive exploration of LLM behaviour during code generation. Users can apply modifications at any point in a generation by replacing a token with an alternative candidate or custom text and continuing generation from the modified prefix, allowing rapid exploration of counterfactual outputs and failure modes. By aligning decoding-time uncertainty and attention signals with code structure via Abstract Syntax Tree (AST) analysis, TokenScope helps practitioners diagnose where confidence degrades, identify structurally fragile regions of generated code, and trace how local token-level decisions propagate into higher-level program behaviour. Together, these capabilities provide a practical environment for debugging, auditing, and understanding LLM-generated code beyond post-hoc inspection.

This paper makes the following contributions: 
\begin{itemize}
    \item we introduce TokenScope, an interactive tool that exposes decoding-time token-level signals for transformer-based LLMs, with a focus on code generation;
    \item we provide integrated support for token-level uncertainty metrics, attention saliency, and interactive branching for counterfactual generation; 
    \item we enable code-aware analysis by aligning token-level signals with AST entities at multiple levels of granularity; and 
    \item we present a unified environment that combines explainability, interpretability, and user-driven exploration, addressing limitations of existing tools that focus on post-hoc analysis or isolated internal signals.
\end{itemize}

\section{Background}
A range of tools has been proposed to improve the interpretability and explainability of transformer-based models. We briefly review the most relevant categories and highlight how TokenScope differs.

General-purpose interpretability platforms such as LIT \cite{lit-tenney2020language} and Phoenix \cite{phoenix} provide interfaces for dataset inspection, model comparison, and embedding analysis. While valuable for evaluation and debugging at the dataset level, these tools do not expose decoding-time token probabilities, alternative candidates, or uncertainty metrics during generation.

Mechanistic interpretability frameworks such as TransformerLens \cite{nanda2022transformerlens} and sparse autoencoder approaches \cite{cooney2023SparseAutoencoder} enable detailed inspection of attention heads, neurons, and internal activations. These approaches offer powerful analytical capabilities but typically require offline analysis and substantial technical expertise, and they are not designed for interactive exploration of individual generations.

Visualization tools such as Ecco \cite{alammar-2021-ecco} and Attention Analysis \cite{bertvis-clark2019what} focus primarily on attention weights. While useful for understanding information flow, attention alone is an incomplete proxy for model confidence or decision-making, and these tools do not integrate uncertainty measures or structural program analysis.


\section{System Overview}
TokenScope is an interactive explainability and interpretability system for decoder-only transformer LLMs, designed to expose fine-grained token-level signals during text and code generation. The system captures internal model statistics at decoding time and presents them through an interactive web interface that supports inspection, comparison, and counterfactual exploration of generation trajectories.

TokenScope follows a modular three-component architecture, enabling support for arbitrary decoder-only models via custom generation server implementations, as illustrated in Figure \ref{fig:system}. 
\begin{itemize}
    \item (i) The generation server is responsible for model discovery, loading, and low-level decoding, exposing Hugging Face–compatible decoder-only models through a minimal generation API. 

    \item (ii) The orchestration server coordinates generation requests through the generation server, manages user sessions, branching generation states, metric computation and aggregation, and frontend-facing API endpoints. 

    \item (iii) The web-based frontend, implemented in React, provides interactive visualizations and controls for exploring model behaviour.
\end{itemize}

\begin{figure}[h]
  \centering
  \includegraphics[width=\linewidth]{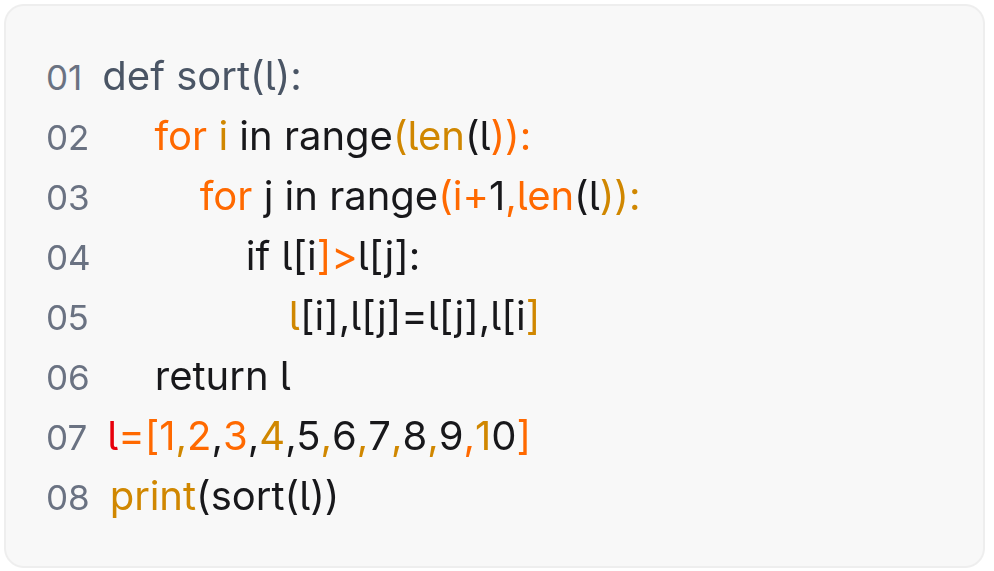}
  \caption{Output generated by Qwen2.5 Coder 1.5B Base using greedy generation, when given `\textcolor{gray}{\texttt{def sort(l):}}' as input. Grey represents the prompt, while other token colors represent token-level confidence scores in an increasing order of Red, Orange, Dark Yellow and Black.}
  \label{fig:generation}
\end{figure}

Generation is performed incrementally and streamed to the frontend. At each decoding step, TokenScope records token probabilities, alternative candidates, attention weights, and derived uncertainty metrics, associates them with the generated token, and makes them available for interactive visualization and analysis (Figure \ref{fig:generation}).

\section{Functionality Overview}
TokenScope is designed as an interactive environment that exposes and manipulates the generation process of decoder-based LLMs. This section provides an overview of its core functionalities.

\subsection{Generation Modes and Decoding Control}
TokenScope supports multiple generation modes, including greedy decoding, temperature-controlled sampling, top-k, and nucleus (top-p) sampling. This enables controlled experiments on how decoding strategies influence reliability and diversity in code generation.

\begin{figure}[h]
  \centering
  \includegraphics[width=\linewidth]{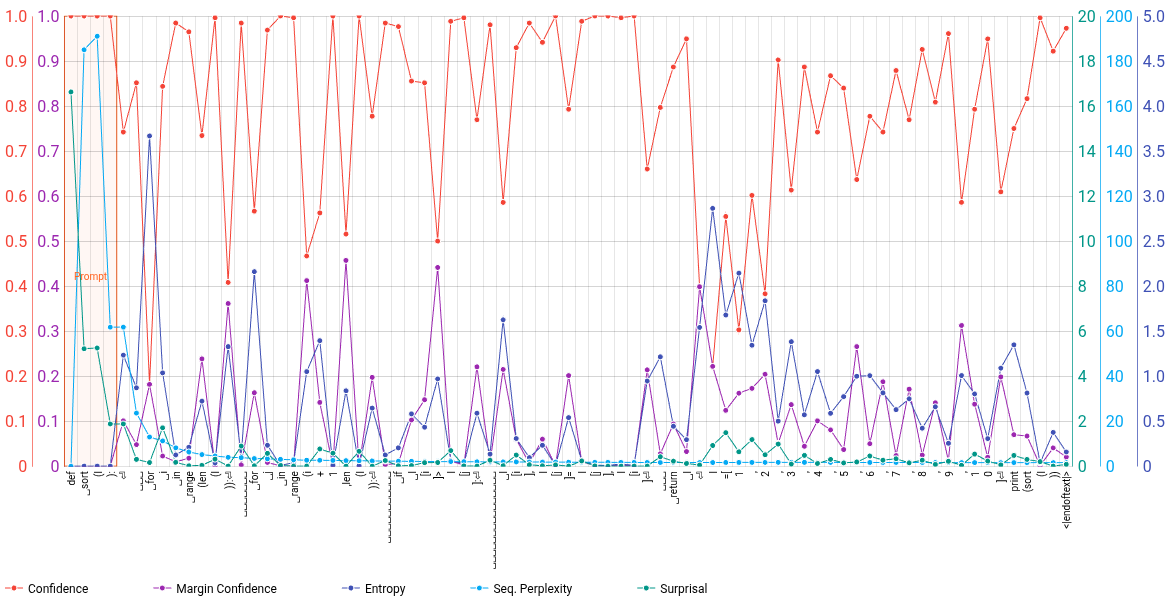}
  \caption{Confidence, Margin Confidence, Entropy, Sequence Perplexity and Surprisal metrics for each token in the output sort function. X and Y axes represent token and metric values, respectively.}
  \label{fig:metrics_chart}
\end{figure}
\subsection{Decoding-time Metrics}
During generation, TokenScope records token-level statistics such as probability, rank among candidates, token surprisal, entropy, and margin confidence, as well as sequence-level metrics such as sequence perplexity. These metrics are displayed as color-coded tokens (Figure \ref{fig:generation}) and plotted across all tokens (Figure \ref{fig:metrics_chart}), and can be aggregated over spans or structural units. Formal definitions and formulas are provided in Section \ref{sec-4}.

\begin{figure}[h]
  \centering
  \includegraphics[width=\linewidth]{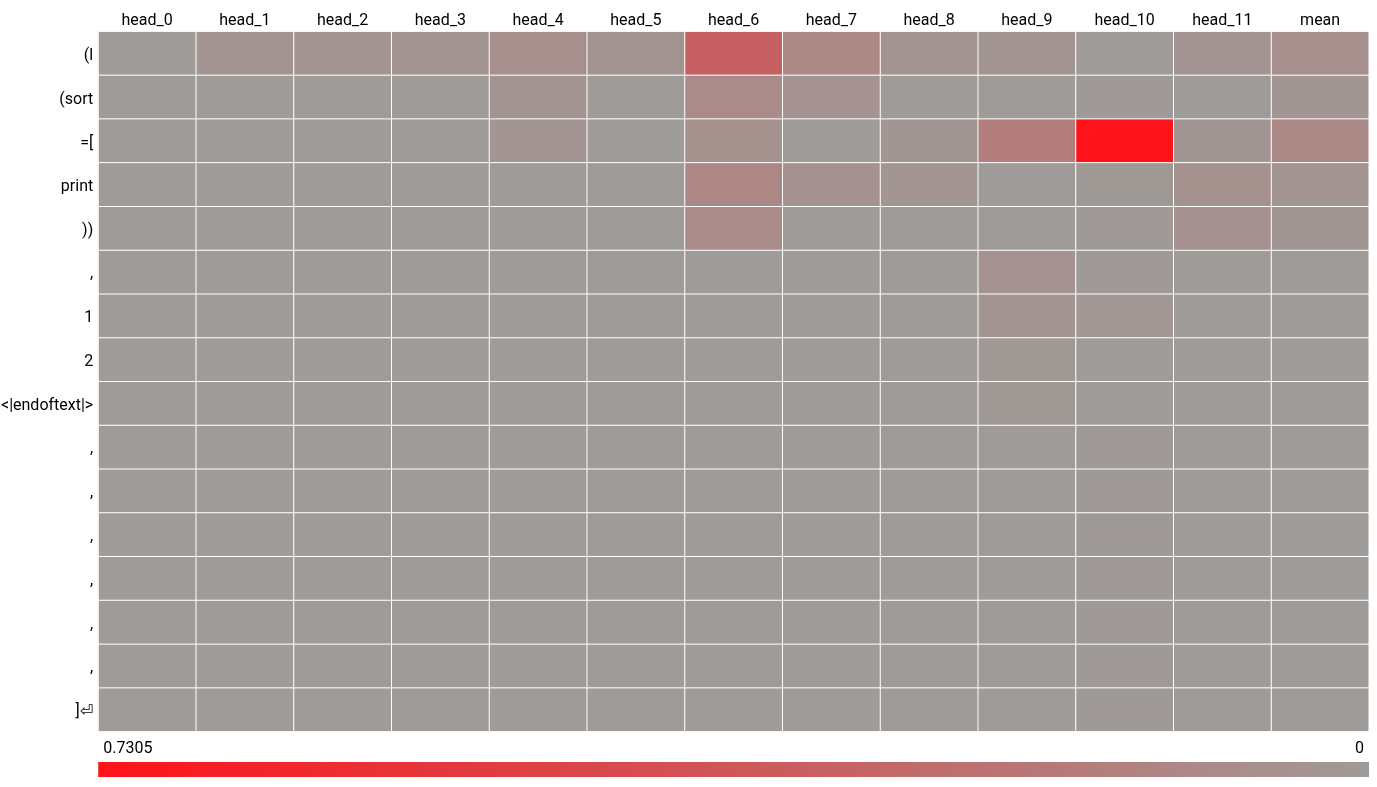}
  \caption{Attention mass of subsequent tokens to token `\emph{\texttt{l}}' (first token on line seven) in the output sort function per attention head and the average of all attention heads.}
  \label{fig:rev-attn}
\end{figure}
\subsection{Attention Analysis}
TokenScope exposes attention weights between each generated token and the prior context at decoding time. From these weights, attention saliency measures how strongly subsequent generated tokens attend to a given token. These signals can be visualized at the token level per each attention head (Figure \ref{fig:rev-attn}) or aggregated over syntax-aware units to analyze long-range dependencies in code generation.

\begin{figure}[h]
  \centering
  \includegraphics[width=0.49\linewidth]{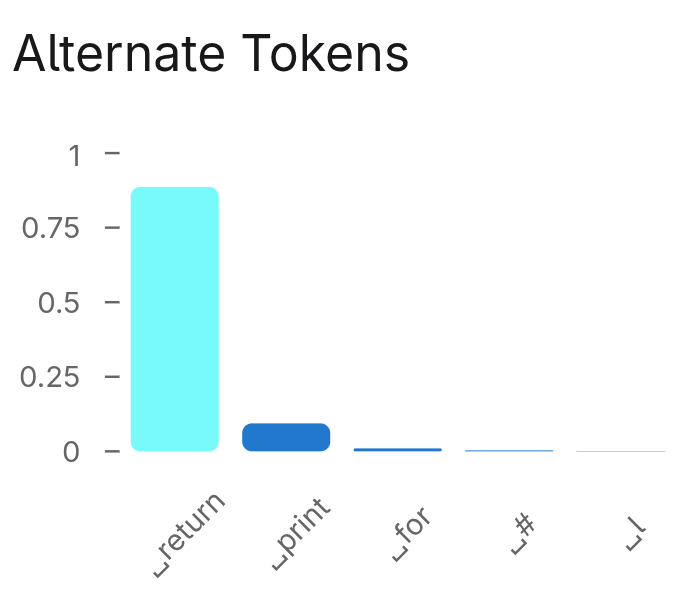}
  \includegraphics[width=0.49\linewidth]{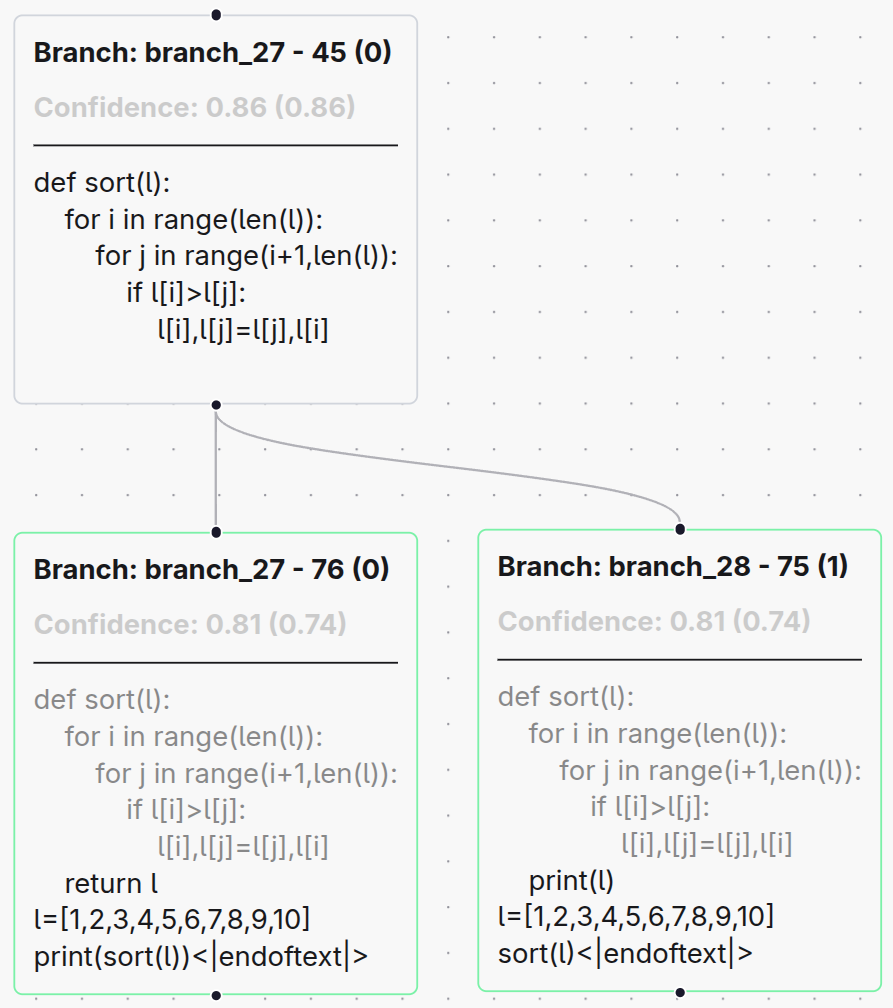}
  \caption{Left: Five top alternative tokens for token `\emph{\texttt{ return}}' in the output sort function. Right: Original output (left leaf node) and the alternative branch when `\emph{\texttt{ return}}' is replaced with token `\emph{\texttt{ print}}'.}
  \label{fig:branching}
\end{figure}
\subsection{Interactive Branching and Counterfactual Generation}
Users can interactively replace any generated token with an alternative candidate (Figure \ref{fig:branching}, left) or custom input and continue generation from the modified prefix. TokenScope maintains a branching structure that records shared prefixes and divergent continuations, enabling systematic exploration of alternative reasoning paths (Figure \ref{fig:branching}, right).

\subsection{Code-Aware Structural Analysis.}
Using Tree-Sitter parsers, TokenScope maps generated tokens to AST nodes such as expressions, statements, and functions. Token-level metrics and attention signals can be aggregated over these structures, allowing analysis of uncertainty and saliency at semantically meaningful levels of abstraction.

\section{Metrics Definitions.} \label{sec-4}
This section defines the core metrics exposed by TokenScope. Let $p_t(v)$ denote the probability assigned by the model to token $v$ at generation step $t$.

\textbf{Token Probability and Rank.}
The probability of the generated token $y_t$ is given by:

\begin{equation}
    p_t(y_t)
\end{equation}

We refer to this probability as the \emph{confidence score} of the token.

Token rank is defined as the position of $y_t$ when candidate tokens are sorted in descending order of probability.

\textbf{Token Surprisal.}
Token surprisal measures the unexpectedness of a generated token:

\begin{equation}
    Surprisal(y_t) = -log p_t(y_t)
\end{equation}

Intuitively, this metric measures how unexpected a generated token is given the preceding context.

\textbf{Margin Confidence.}
Margin confidence captures the difference between the most likely and the second most likely tokens. This metric reflects local decisiveness rather than absolute correctness.



\textbf{Entropy.}
Token-level entropy quantifies uncertainty over the candidate distribution:

\begin{equation}
    H_t = -\sum_{v}p_t(v)log p_t(v)
\end{equation}

\textbf{Sequence Perplexity.}
Sequence perplexity over a generated sequence $y_{1:T}$ is defined as

\begin{equation}
    Perplexity = exp(\frac{1}{T - 1}\sum_{t=2}^T - log p_t(y_T))
\end{equation}

which quantifies the average unpredictability of the full generated sequence, where lower values indicate the model finds the sequence more probable.

\textbf{Attention Mass.}
For a generated token $t_i$, attention mass is defined as the average amount of attention it receives from all subsequently generated tokens $t_j$ where $j > i$. Formally, for each attention head $h$,

\begin{equation}
    AM_h(t_i) = \frac{1}{N-1}\sum_{j=i+1}^N A_h(j,i),
\end{equation}

where $A_h(j,i)$ denotes the attention weight from token $t_j$ to token $t_i$ in head $h$.

Intuitively, attention mass measures how much a token continues to influence later decoding decisions.

\section{Code-Aware Structural Analysis}
In Code Analysis Mode, TokenScope aligns generated tokens with AST entities using Tree-Sitter parsers.

\subsection{Token–Entity Alignment and Aggregation}

LLM token boundaries do not generally align with AST entities, as tokenization is independent of syntactic parsing. An AST entity is considered a candidate for an LLM token if the token is fully contained within the entity span, overlaps with the entity start, or overlaps with the entity end.

When multiple entities are candidates for a single token, TokenScope assigns the entity with the highest predefined priority, reflecting semantic importance (e.g., identifiers over delimiters). 




\subsection{Code Analysis Modes}
To support analysis at different semantic resolutions, the tool provides five AST entity modes, each defining a distinct aggregation level for token-level metrics and attention signals.


\textbf{Token Mode} represents the finest granularity, where each generated token is analyzed independently. This mode is useful for inspecting localized uncertainty, abrupt probability drops, or attention shifts at specific lexical positions.

\textbf{Expression Mode} groups tokens into syntactic expressions such as literals, binary operations, function calls, or indexing expressions. Aggregating metrics at this level highlights uncertainty and attention patterns within semantic units that often correspond to logical decisions in code.

\textbf{Statement Mode} aggregates tokens belonging to complete statements, including assignments, control-flow statements, and return statements. This mode enables comparison of confidence and attention across executable units that affect program behaviour.

\textbf{Line Mode} aggregates tokens belonging to a single line of code, as determined by source-level line boundaries after formatting. This mode enables analysis of uncertainty and attention at a granularity commonly used by developers when reading and debugging code, and supports identifying lines that disproportionately contribute to low-confidence generations or error propagation.

\textbf{Block Mode} groups tokens within structural blocks such as function bodies, loops, or conditional branches. Analysis at this level supports studying how uncertainty and attention evolve across larger scopes and control-flow regions.

Together, these five modes allow TokenScope users to move seamlessly between fine-grained token inspection and higher-level structural analysis, supporting both detailed debugging and holistic evaluation of generated code.

\section{Limitations}
TokenScope currently targets decoder-only autoregressive models and requires access to token probabilities and attention weights, which limits its applicability to closed or heavily abstracted LLM APIs. Its code-aware analysis relies on Tree-Sitter parsers and is presently limited to Python; while support for additional languages is feasible, structural analysis depends on parser availability and may fail on incomplete or syntactically invalid generations. Finally, TokenScope is designed as an evaluation and analysis tool rather than a production inference system, and the overhead of extracting, aggregating, and transferring fine-grained interpretability signals prevents it from matching the performance of deployment-oriented inference frameworks.

\section{Conclusion}
We presented TokenScope, an interactive tool for observing, analyzing, and manipulating token-level behaviour in decoder-based LLMs for code generation. By exposing decoding-time uncertainty metrics, attention signals, interactive branching, and code-aware structural aggregation, TokenScope enables analyses that are difficult to perform with existing tools. TokenScope can support both practical debugging and empirical research on LLM behaviour in software engineering. TokenScope's source code is open in order to facilitate further study in this direction.

\section*{Data Availability}
TokenScope's source code is open and available on Github\footnote{\url{https://github.com/Amirresm/tokenscope}}.




\bibliographystyle{ACM-Reference-Format}
\bibliography{bib}

\clearpage
\onecolumn
\appendix

\section{Tool Availability}
TokenScope is fully open-source. The source code is publicly available at \url{https://github.com/Amirresm/tokenscope}. To support reproducibility and ease of deployment, the repository includes a \emph{Dockerfile} that enables consistent setup across environments.

\section{Screencast}
A screencast demonstrating TokenScope is available at \url{https://youtu.be/HI1L1X9LruQ}. This video is also available for download \footnote{\url{https://raw.githubusercontent.com/Amirresm/tokenscope/main/supplementary/demo/demo.mp4}}.

\section{Tool Walkthrough}
This section provides an illustrative walkthrough of TokenScope for a simple code generation scenario. In this scenario, the base variant of Qwen 3 0.6B model is used to continue the input \texttt{def sort(l):}, effectively generating a sort function in a code completion setting.

\begin{figure}[h]
  \centering
  \includegraphics[width=\linewidth]{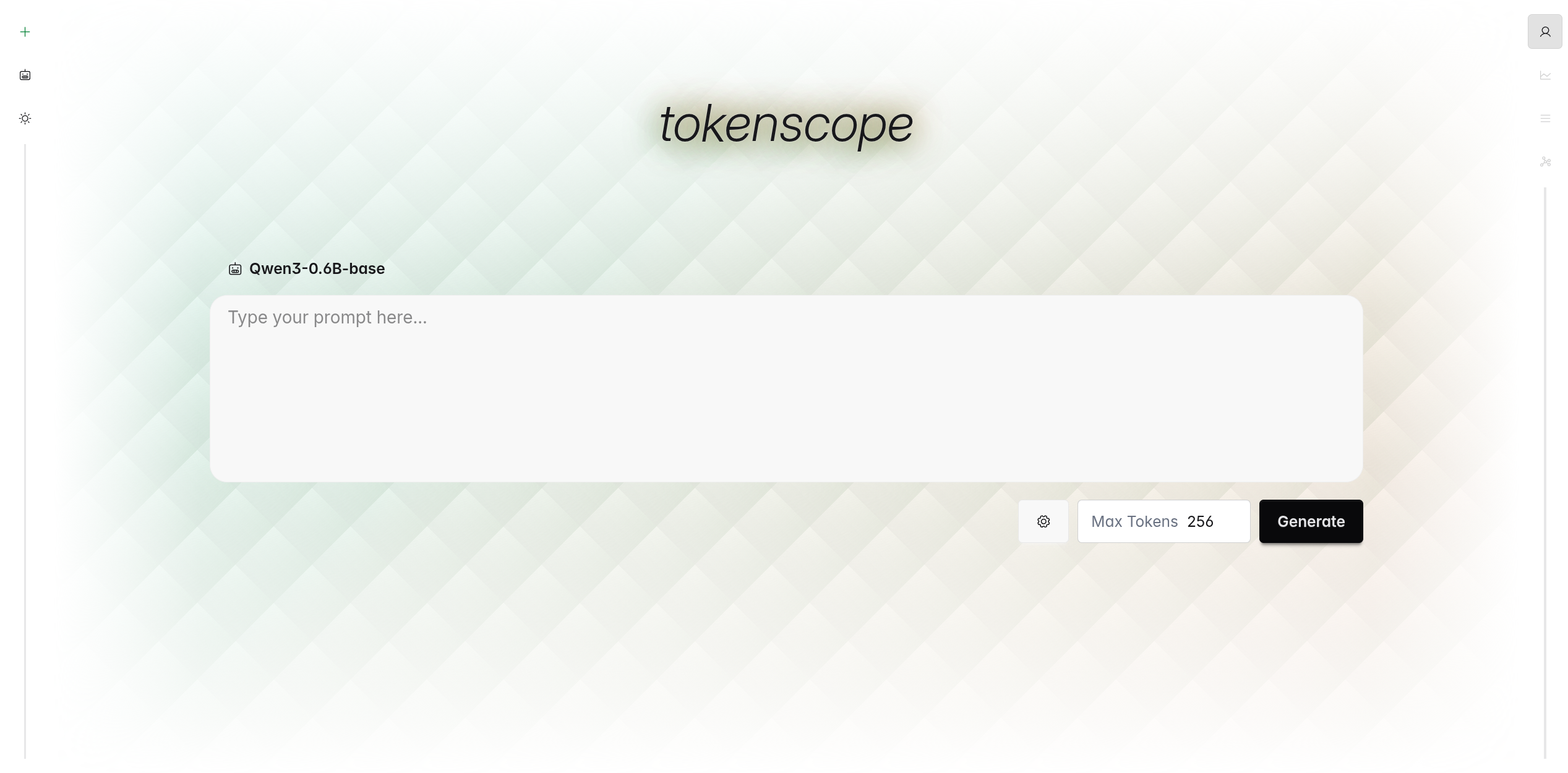}
  \caption{Users choose the model, prompt and generation settings on the landing page.}
  \label{fig:wt-home}
\end{figure}

\subsection{Landing Page}

\begin{figure}[h]
  \centering
  \includegraphics[width=0.49\linewidth]{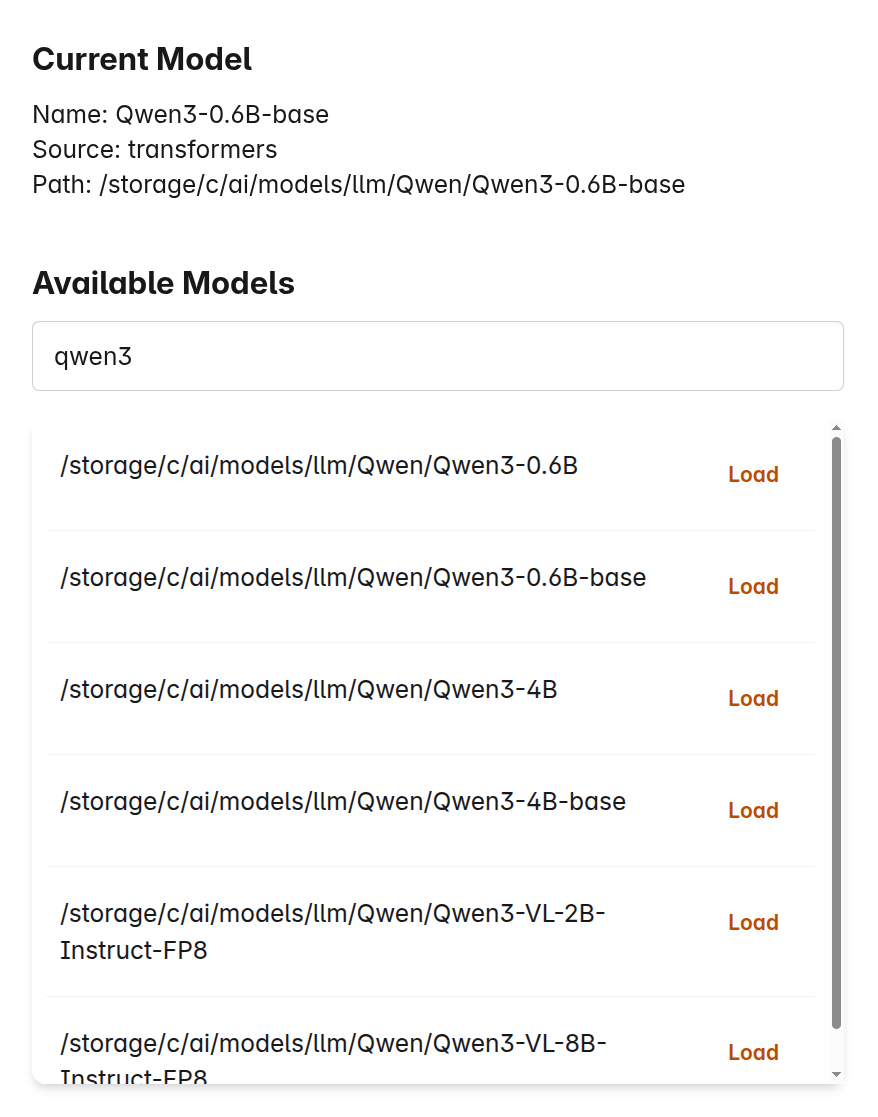}
  \includegraphics[width=0.49\linewidth]{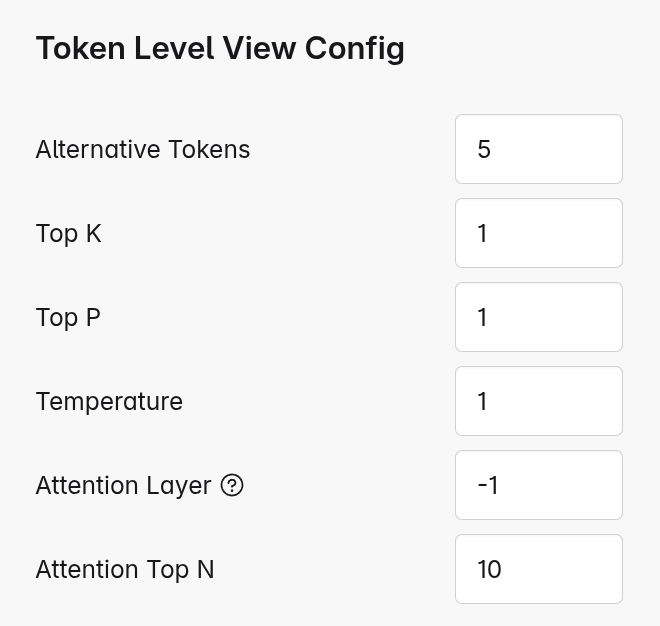}
  \caption{Left: The model management modal displays the currently loaded and all available models. Right: The generation settings drop-down allows defining the number of alternative tokens, sampling settings and attention configuration.}
  \label{fig:wt-llm-modal-gen-settings}
\end{figure}


Figure \ref{fig:wt-home} presents the home page. User starts by loading a model (left on Figure \ref{fig:wt-llm-modal-gen-settings}), entering a prompt, configuring the generation settings (right on Figure \ref{fig:wt-llm-modal-gen-settings}) and selecting the \textbf{generate} button.

\begin{figure}[h]
  \centering
  \includegraphics[width=\linewidth]{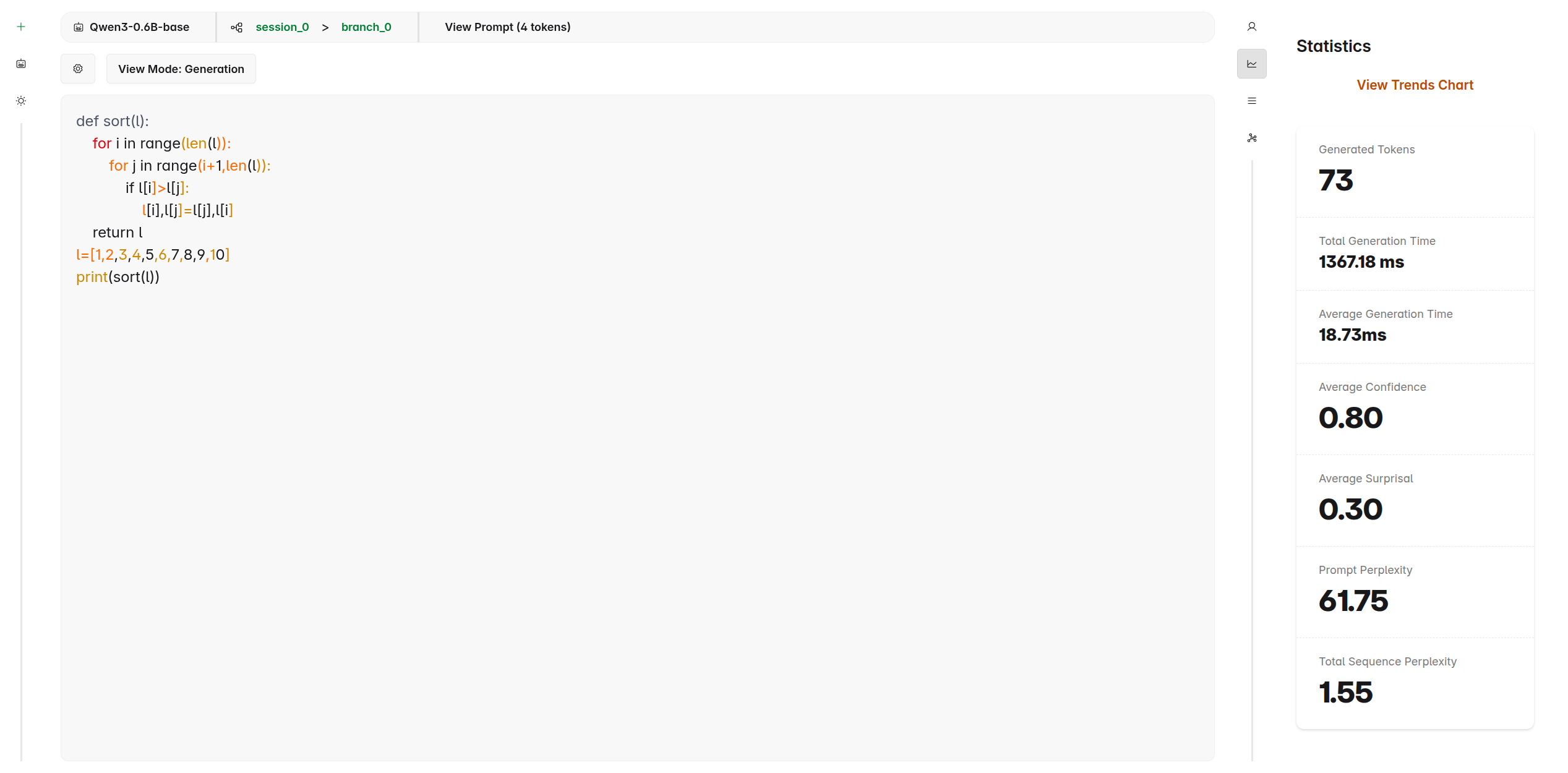}
  \caption{The main content view displays the generated output and a high-level summary of the current generation metrics and metadata.}
  \label{fig:wt-gen}
\end{figure}

\subsection{Main Content View}
When generation starts, the user is redirected to the main content view (Figure \ref{fig:wt-gen}), where the generated output and a high-level summary of the current generation metrics and metadata are visible.

\begin{figure}[h]
  \centering
  \includegraphics[width=\linewidth]{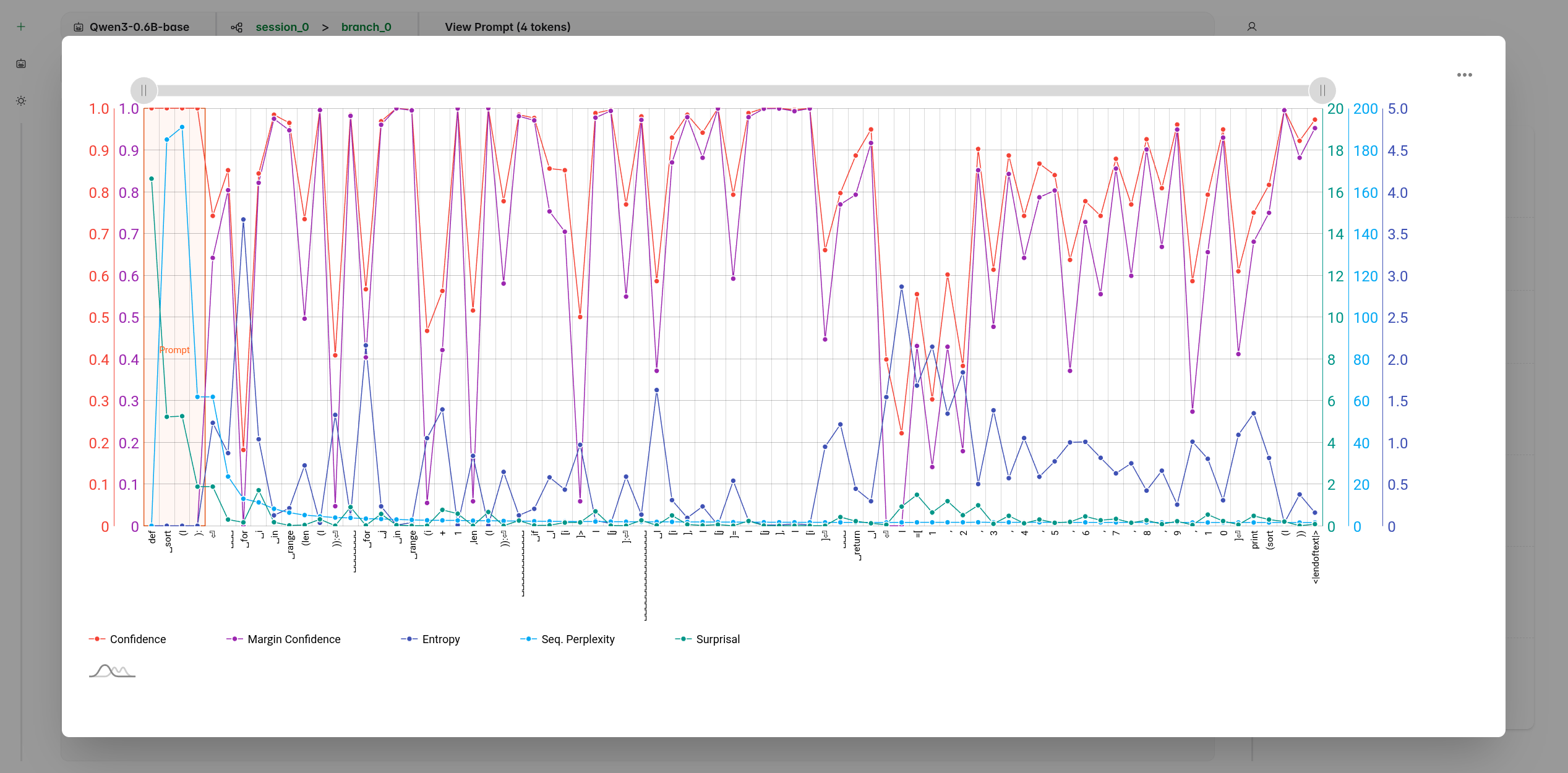}
  \caption{Trends chart displays an overview of all metrics for each token across the generation.}
  \label{fig:wt-trends}
\end{figure}

Selecting the \textbf{View Trends Chart} on the \emph{Statistics} sidebar opens the trends chart (Figure \ref{fig:wt-trends}), which provides an overview of all metrics per token across the generation.

\begin{figure}[h]
  \centering
  \includegraphics[width=\linewidth]{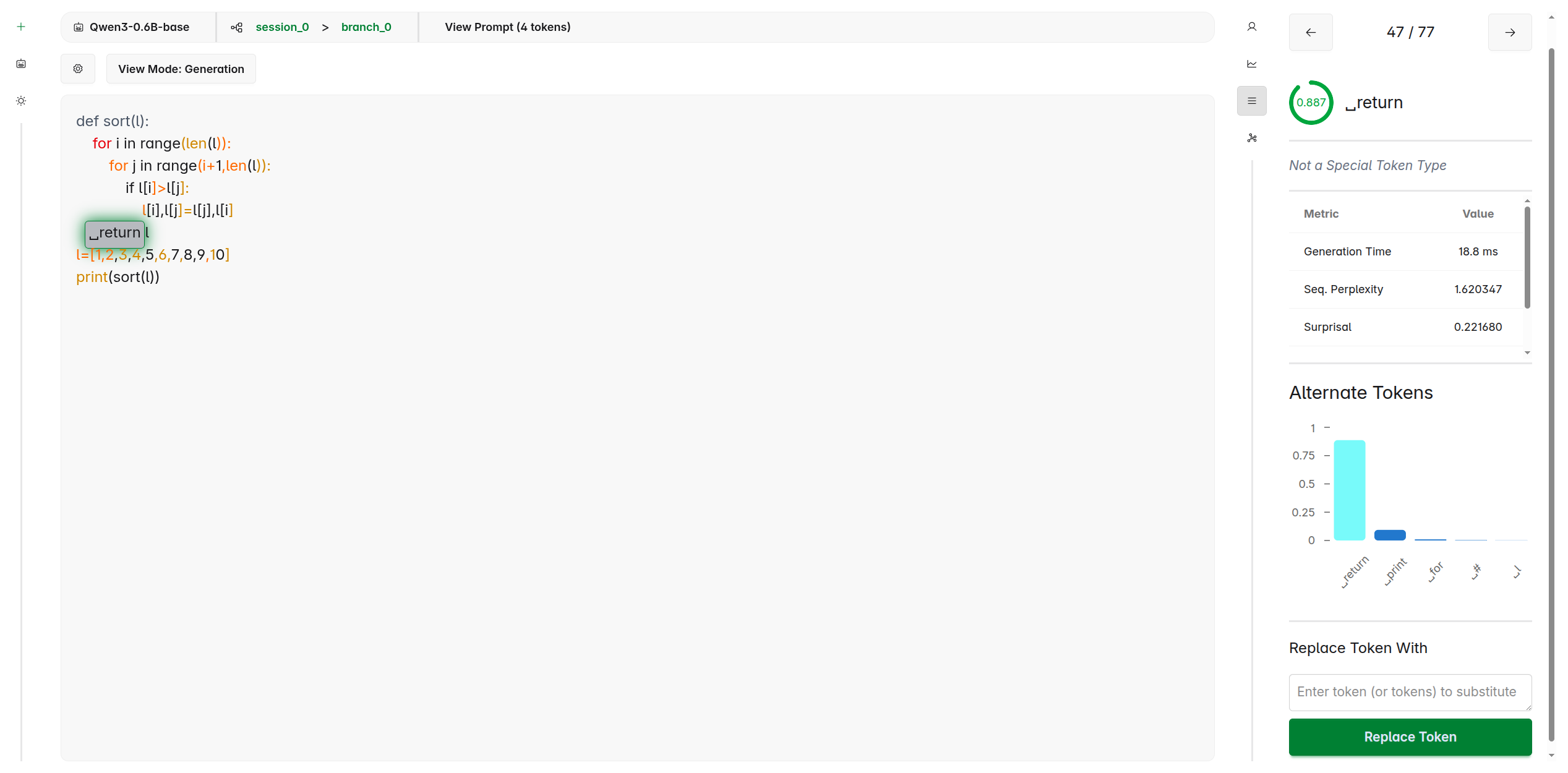}
  \caption{The Token Details sidebar presents special token types, token metrics and alternative tokens, and allows for creating new branches in the generation.}
  \label{fig:wt-token-details}
\end{figure}

Selecting an individual token switches to the \textbf{Token Details} sidebar (Figure \ref{fig:wt-token-details}), presenting the confidence score and the token's string (with whitespace characters such as new line visualized), special token types, token metrics and alternative tokens.

\begin{figure}[h]
  \centering
  \includegraphics[width=\linewidth]{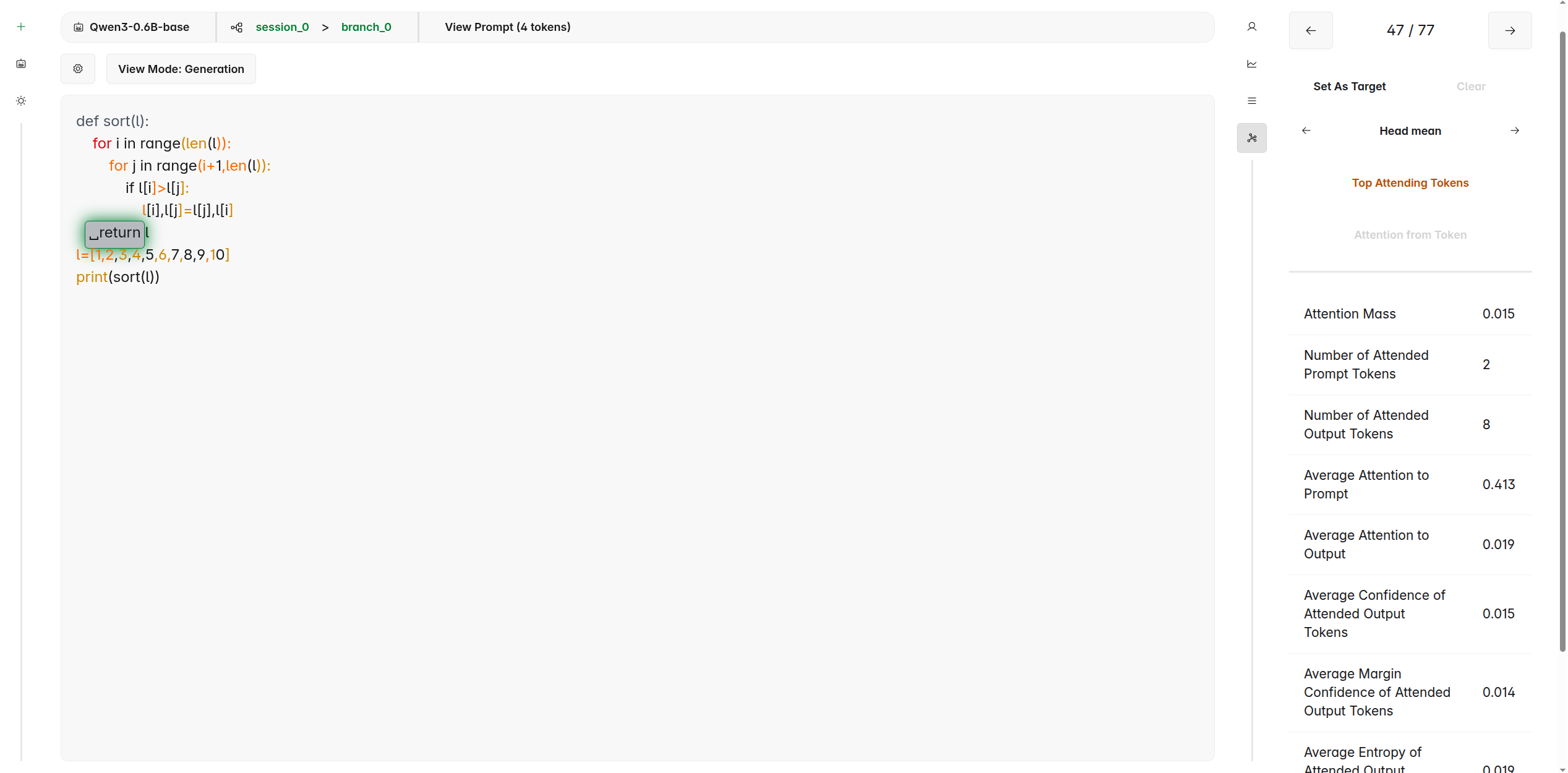}
  \caption{The Attention Details sidebar displays tokens' attention-relevant metrics, attention source tokens and subsequent attending tokens.}
  \label{fig:wt-attn}
\end{figure}

\begin{figure}[h]
  \centering
  \includegraphics[width=\linewidth]{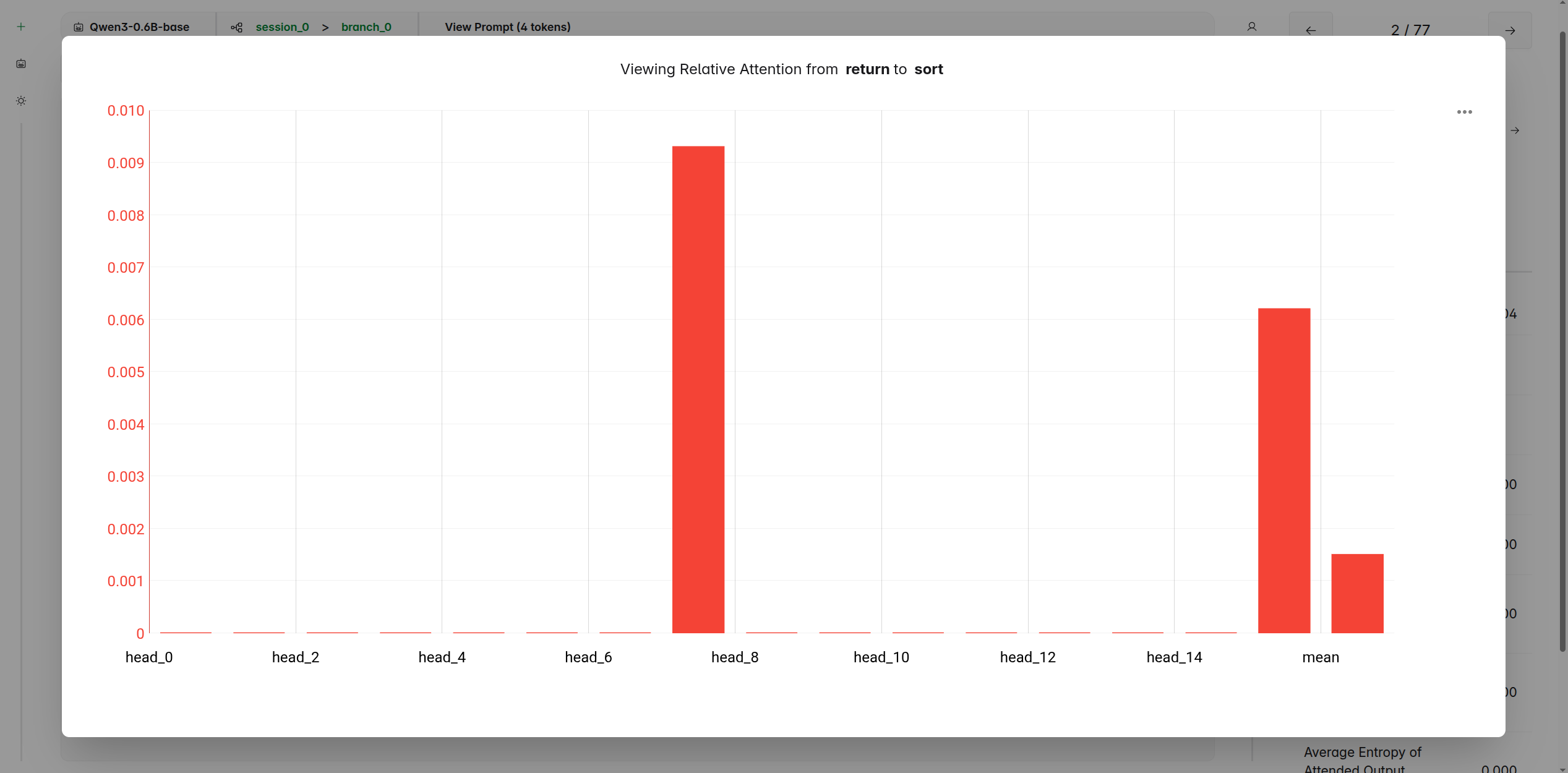}
  \caption{When a target attention token is selected, the user can plot the attention amount of any attention source token for each head.}
  \label{fig:wt-attn-bar}
\end{figure}

\begin{figure}[h]
  \centering
  \includegraphics[width=\linewidth]{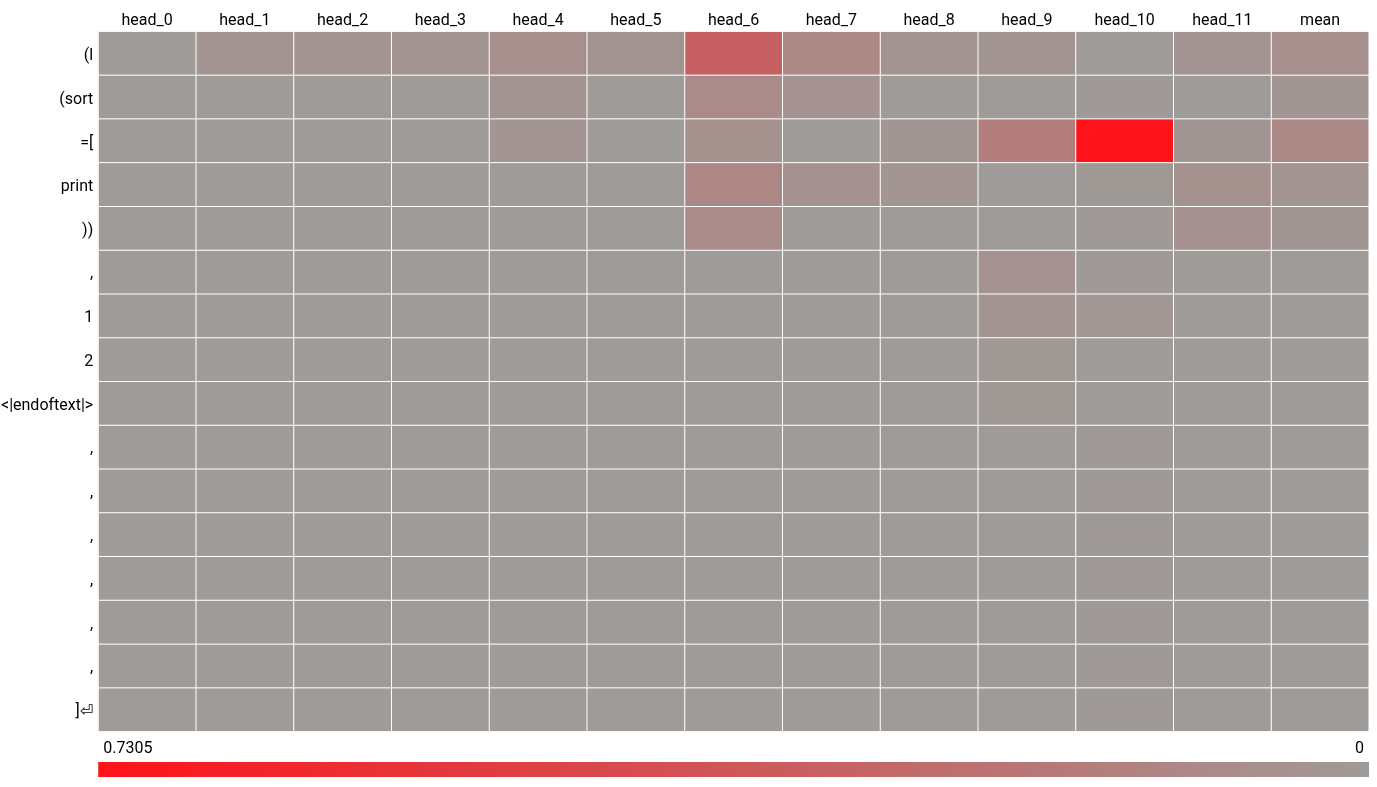}
  \caption{Heatmap plot of subsequent attending tokens per attention head for the target token.}
  \label{fig:wt-rev-attn}
\end{figure}

With an individual token selected, the user can switch to the \textbf{Attention Details} sidebar (Figure \ref{fig:wt-attn}) to examine the token's attention-relevant metrics such as \emph{attention mass}, attention source tokens (Figure \ref{fig:wt-attn-bar}) and subsequent attending tokens (Figure \ref{fig:wt-rev-attn}).

\begin{figure}[h]
  \centering
  \includegraphics[width=\linewidth]{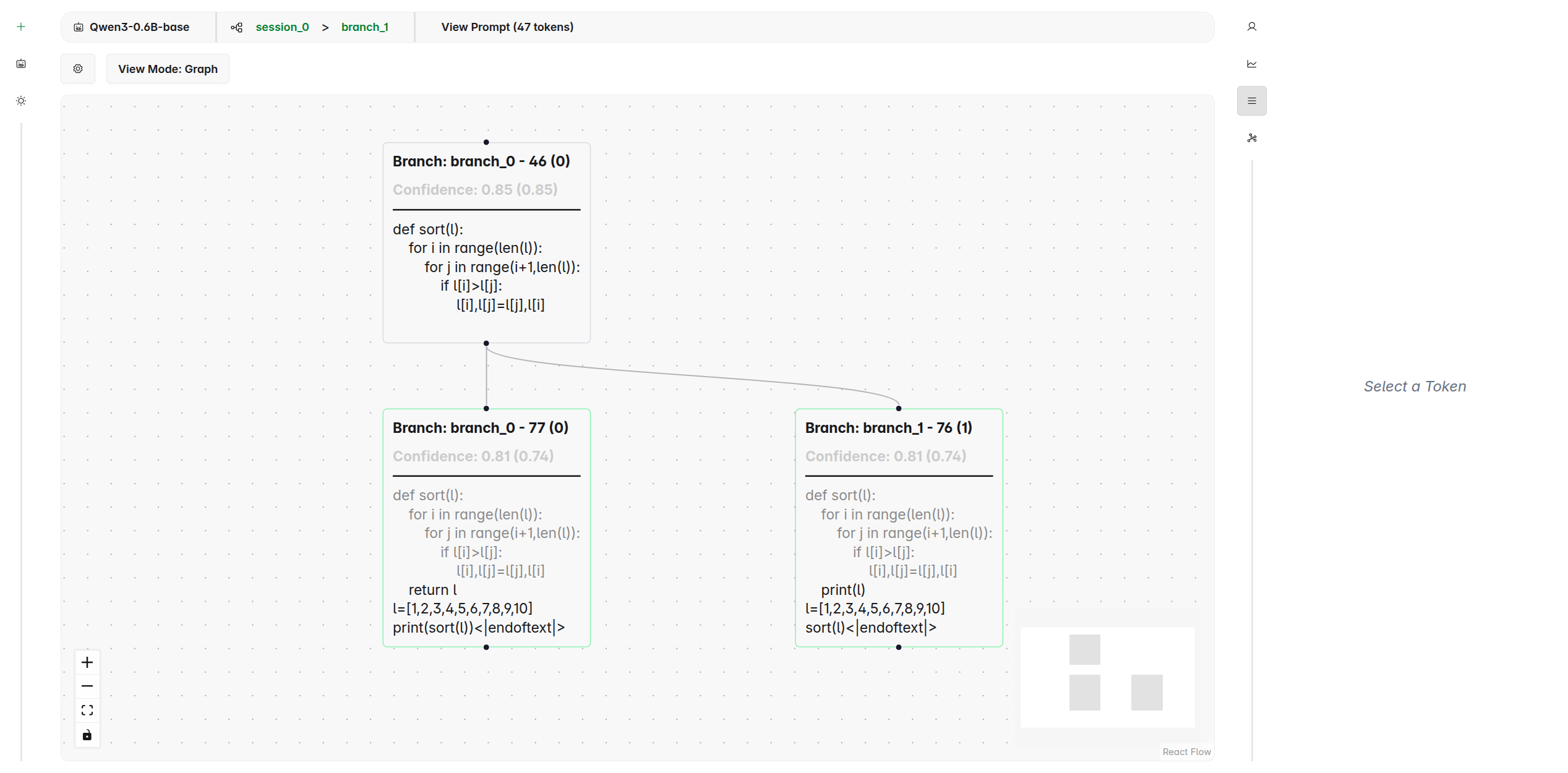}
  \caption{Graph view provides an overview of all branches of the generation and a summary of each branch.}
  \label{fig:wt-branch}
\end{figure}

\subsection{Generation Branching}
When an alternative token is selected in the \emph{Token Details} sidebar, a branch in the generation is created, and the generation is continued with the new prefix, which is managed in the \textbf{Graph View}, as illustrated in Figure \ref{fig:wt-branch}.

\begin{figure}[h]
  \centering
  \includegraphics[width=\linewidth]{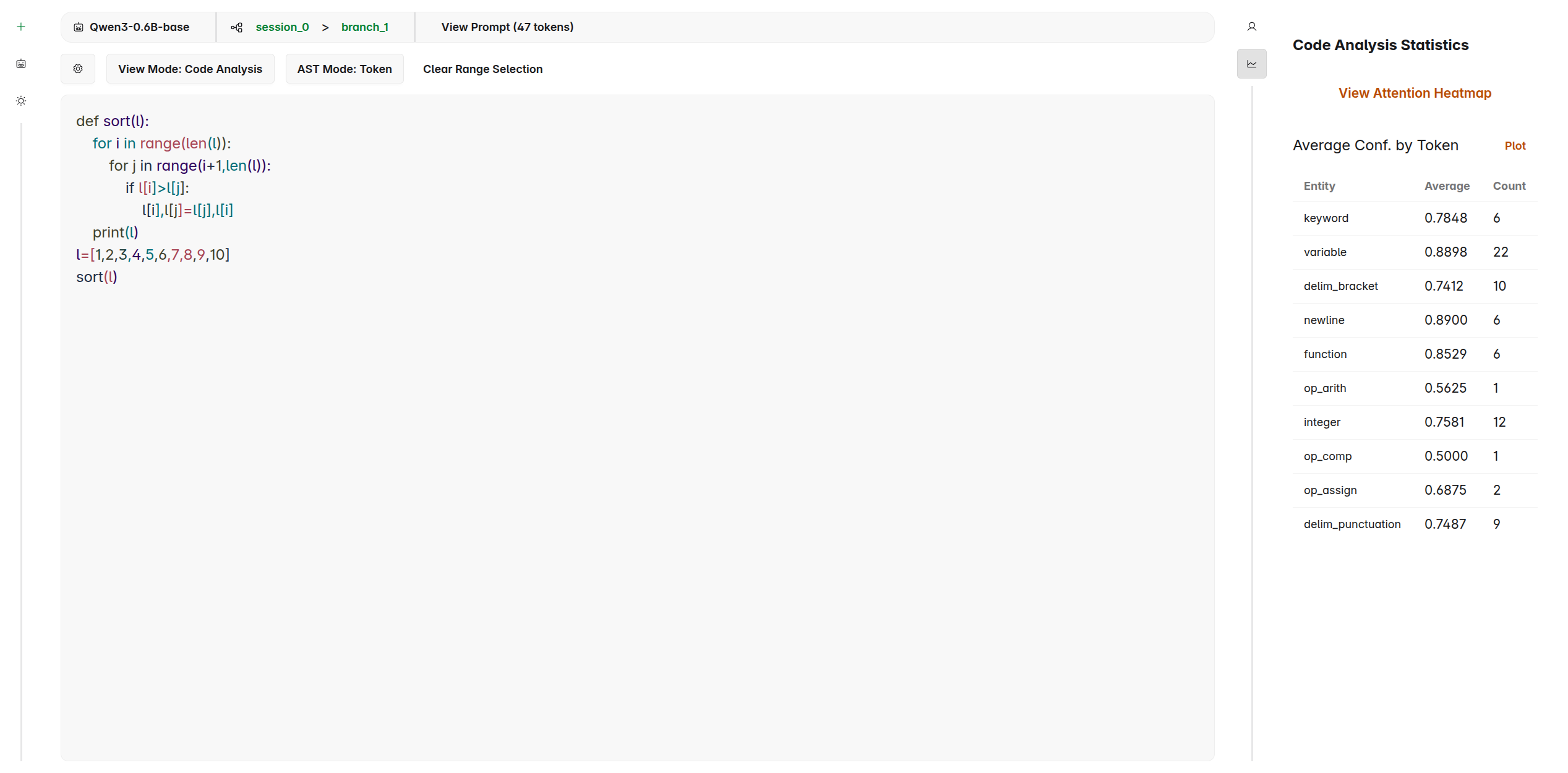}
  \caption{Code Analysis view displays extracted code entities and their average confidence scores.}
  \label{fig:wt-ast}
\end{figure}

\begin{figure}[h]
  \centering
  \includegraphics[width=\linewidth]{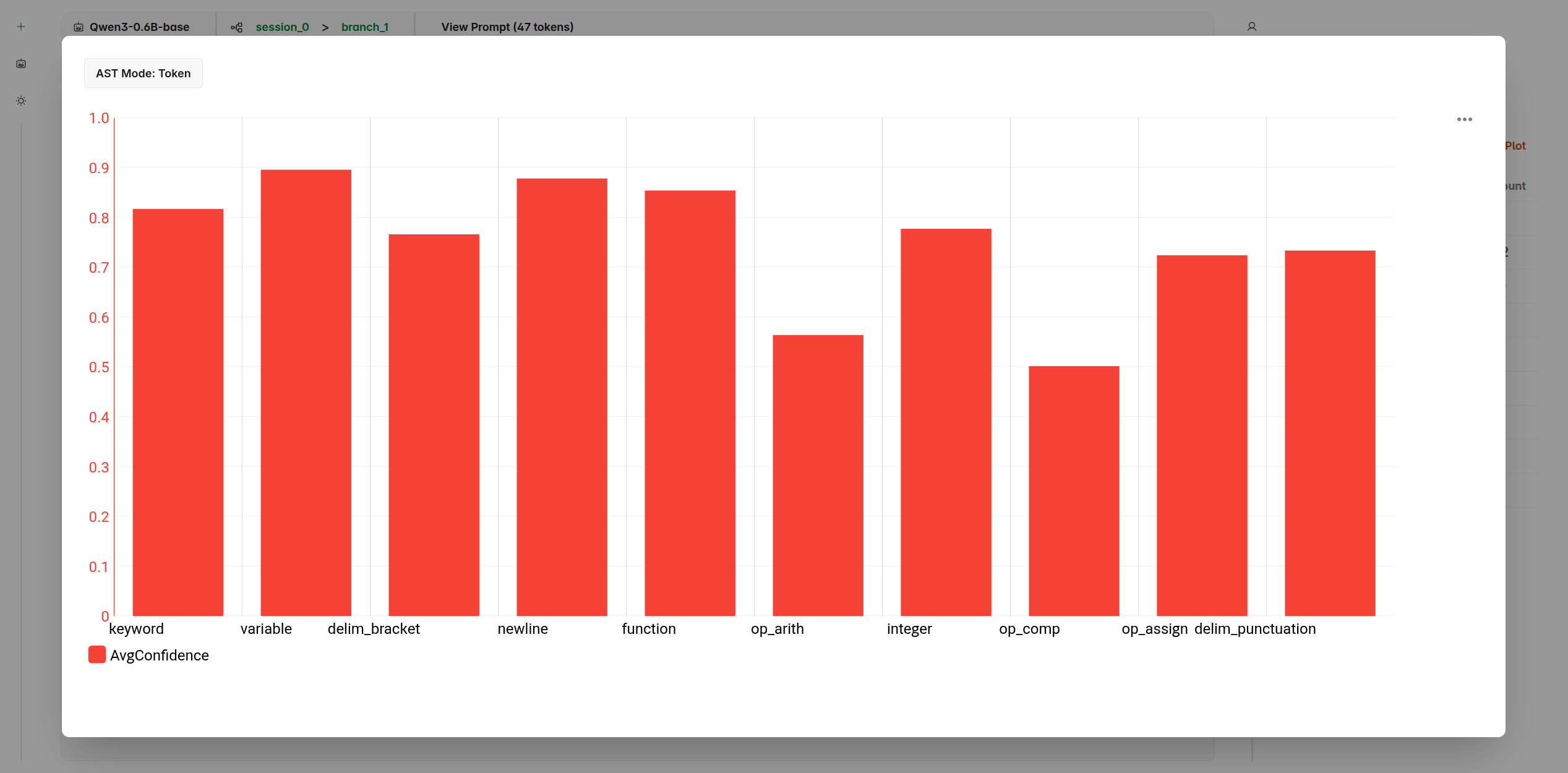}
  \caption{Plot view of the average confidence scores across the extracted code entities.}
  \label{fig:wt-ast-bar}
\end{figure}

\begin{figure}[h]
  \centering
  \includegraphics[width=\linewidth]{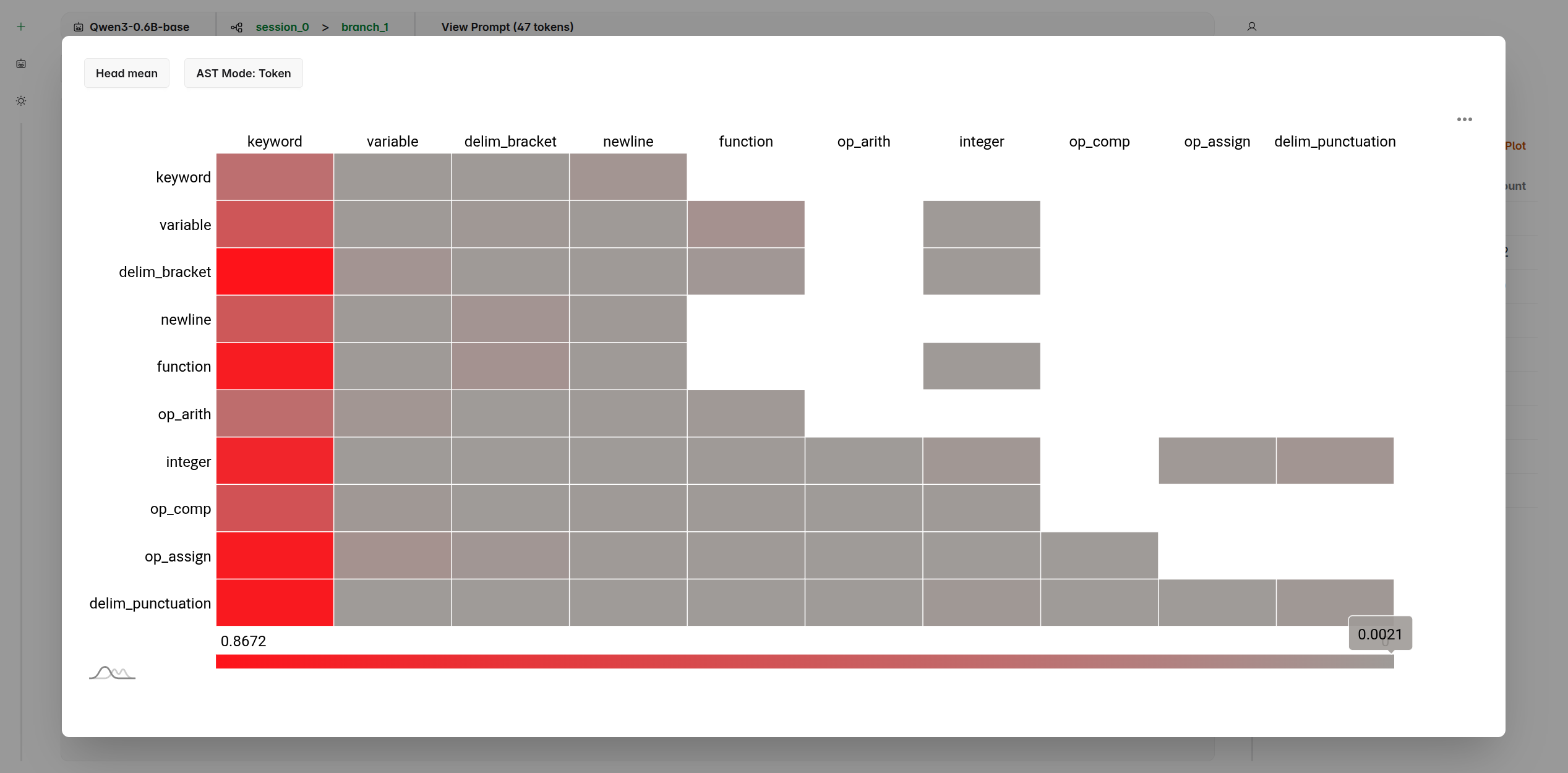}
  \caption{Heatmap of attention relationship between code entities for each attention head.}
  \label{fig:wt-ast-attn}
\end{figure}

\subsection{Code Analysis}
After selecting a region of the generation corresponding to a code snippet, TokenScope analyses, extracts, and assigns code entities to the generated tokens. These entities are presented in the \textbf{Code Analysis} view, shown in Figure \ref{fig:wt-ast}. The \textbf{Code Analysis Statistic} sidebar presents a summary of extracted entities and their average confidence scores. This summary is also provided in a bar plot format (Figure \ref{fig:wt-ast-bar}).

Additionally, TokenScope extracts attention relationships between code entities and presents them as a heatmap, as shown in Figure \ref{fig:wt-ast-attn}.


\end{document}